\documentclass[letterpaper, 10 pt, conference]{ieeeconf}  % Comment this line out if you need a4paper

\IEEEoverridecommandlockouts                              % This command is only needed if 
\usepackage{amsmath} % assumes amsmath package installed
\usepackage{amssymb}  % assumes amsmath package installed
\usepackage{graphicx}
\usepackage{caption}
\usepackage{booktabs}
\usepackage{float}
\usepackage{cite}
\usepackage{color}
\usepackage[table]{xcolor}
\usepackage{multirow}
\usepackage{flushend}
\usepackage{fontawesome5}
\usepackage{hyperref}
\usepackage{fancyhdr}
\fancypagestyle{withfooter}{
  
  \fancyhead[L]{}
  \fancyhead[R]{}
  \fancyfoot[C]{\footnotesize Presented at the 2025 IEEE ICRA Workshop on Field Robotics}
}

\title{\LARGE \bf
VertiSelector: Automatic Curriculum Learning for \\Wheeled Mobility on Vertically Challenging Terrain
}

\author{Tong Xu, Chenhui Pan, and Xuesu Xiao% <-this % stops a space
\thanks{All authors are with the Department of Computer Science, George Mason University {\tt\small \{txu25, cpan7, xiao\}@gmu.edu}}% <-this % stops a space
}

\begin{document}

\maketitle
\thispagestyle{empty}
\pagestyle{empty}
% \thispagestyle{withfooter}
% \pagestyle{withfooter}

%%%%%%%%%%%%%%%%%%%%%%%%%%%%%%%%%%%%%%%%%%%%%%%%%%%%%%%%%%%%%%%%%%%%%%%%%%%%%%%%
\begin{abstract}

Reinforcement Learning (RL) has the potential to enable extreme off-road mobility by circumventing complex kinodynamic modeling, planning, and control by simulated end-to-end trial-and-error learning experiences. However, most RL methods are sample-inefficient when training in a large amount of manually designed simulation environments and struggle at generalizing to the real world. To address these issues, we introduce \textit{VertiSelector} (VS), an automatic curriculum learning framework designed to enhance learning efficiency and generalization by selectively sampling training terrain. VS prioritizes vertically challenging terrain with higher Temporal Difference (TD) errors when revisited, thereby allowing robots to learn at the edge of their evolving capabilities. By dynamically adjusting the sampling focus, VS significantly boosts sample efficiency and generalization within the \texttt{VW-Chrono}\footnote{\faGithub \url{https://github.com/RobotiXX/VertiSelector}} simulator built on the Chrono multi-physics engine. Furthermore, we provide simulation and physical results using VS on a Verti-4-Wheeler platform. These results demonstrate that VS can achieve {23.08}\% improvement in terms of success rate by efficiently sampling during training and robustly generalizing to the real world.

\end{abstract}
%%%%%%%%%%%%%%%%%%%%%%%%%%%%%%%%%%%%%%%%%%%%%%%%%%%%%%%%%%%%%%%%%%%%%%%%%%%%%%%%
\section{Introduction}

Autonomous mobile robots are increasingly being deployed in unstructured, off-road environments for applications such as search and rescue~\cite{xiao2015locomotive, xiao2017uav, murphy2016two}, planetary exploration~\cite{tiwari2019unified, tiwari2018estimating, seeni2010robot}, and agricultural operations~\cite{kumar2020review}. However, navigating extreme terrain with dense and high vertical protrusions from the ground remains a significant challenge~\cite{wermelinger2016navigation}. Traditional approaches rely on sophisticated kinodynamic modeling, motion planning, and vehicle control, which can cause cascading errors and are difficult to develop and adapt to changing conditions~\cite{hutter2016anymal}.

\begin{figure}[h]
    \centering
    \includegraphics[width=\columnwidth]{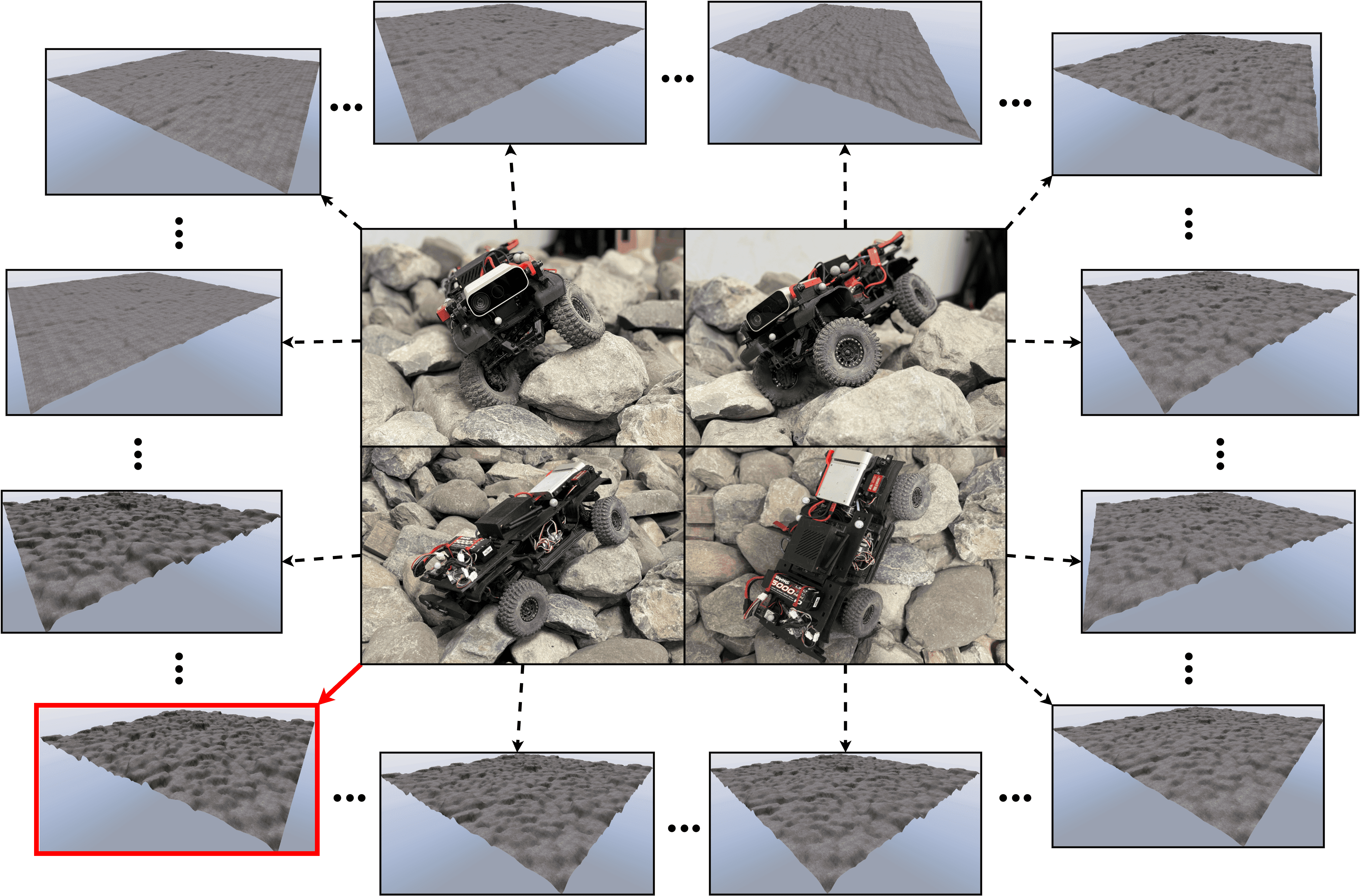}
    \caption{VertiSelector can selectively sample training terrain based on future learning potential in the \texttt{VW-Chrono} simulator to improve RL sample efficiency and generalization.} 
    \label{VW-chrono}
\end{figure}

Reinforcement learning (RL) offers a promising alternative by enabling robots to learn end-to-end motion policies directly from simulated trial-and-error experiences~\cite{kober2013reinforcement}. By circumventing the need for explicit modeling, planning, and control, RL has the potential to achieve more robust and adaptive off-road navigation. Learning from a high-precision physics model in a simulator with RL in advance can also alleviate simulation-to-reality (sim2real) gap during deployment. 

However, RL training requires a large amount of simulation data and can be sample-inefficient. It also often struggles with overfitting to the specific experiences encountered during training, which can significantly limit its ability to generalize to novel situations and hinder its broad applicability~\cite{cobbe2019quantifying}. To address these limitations, Procedural Content Generation (PCG) has emerged as a promising approach~\cite{cobbe2020leveraging, risi2020increasing, jiang2021prioritized}. PCG can algorithmically generate varied configurations before each training episode by modifying the training environments. 
% By associating each generated environment with a unique identifier, such as a fixed index, 
Diverse PCG environments can improve a trained policy's generalization on previously unseen environments and can potentially form a consistent curriculum based on the RL agent's evolving capability. 
% Vertically challenging terrain with various difficulty levels can be generated by PCG. 

To push the boundaries of off-road  wheeled mobility on vertically challenging terrain, we develop a novel Automatic Curriculum Learning (ACL) method, \textbf{V}erti\textbf{S}elector (VS), which leverages differences in learning potential across various terrain produced by PCG to enhance both sample efficiency and generalization of RL. VS works in a set of PCG enviornments in a simulator, \texttt{VW-Chrono}, within the Chrono multi-physics simulation engine~\cite{tasora2016chrono}. Throughout the training process, VS continuously assesses and updates scores that gauge the RL agent's learning potential on each terrain, taking into account its evolving capability and the Temporal Difference (TD) errors observed from the most recent trajectory sampled from that specific terrain.
RL policies efficiently learned with VS in \texttt{VW-Chrono} for navigating vertically challenging terrain can then be deployed onto a physical Verti-4-Wheeler (V4W) platform~\cite{datar2023toward}, showing superior real-world generalizability. 
In summary, the contributions of this work are threefold:

\begin{figure}
    \centering
    \includegraphics[width=\columnwidth]{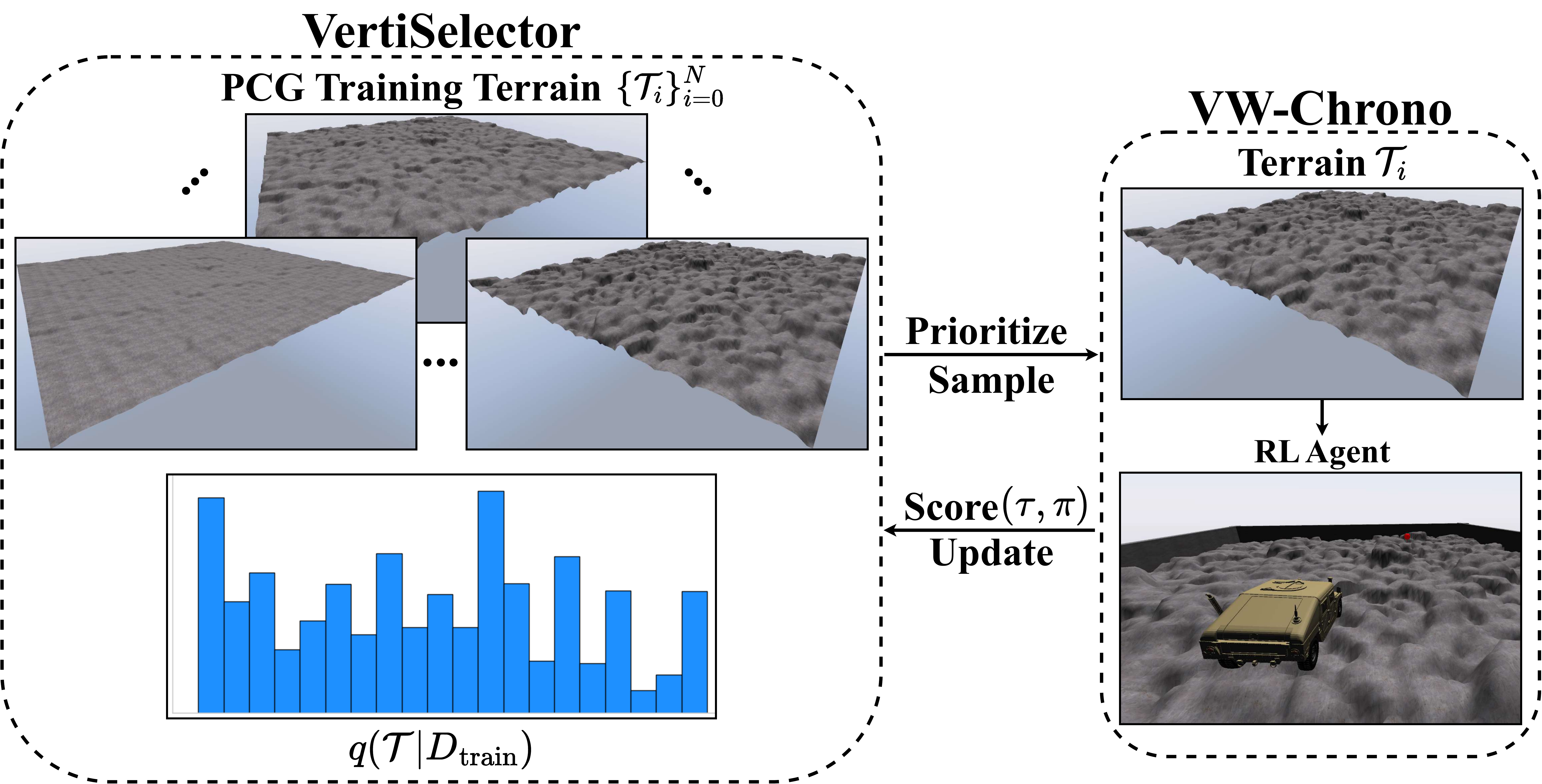}
    \caption{VertiSelector Overview: The next training terrain $\mathcal{T}_i$ is  sampled from the training distribution over the PCG-produced terrain set based on priorities determined by evaluation scores of the current policy $\pi$. A trajectory $\tau$ on this terrain $\mathcal{T}_i$ is used to update the training distribution.}
    \label{VS}
\end{figure}

\begin{itemize}
    \item We present the \texttt{VW-Chrono} simulator (Fig.~\ref{VW-chrono}), designed for wheeled mobility on vertically challenging terrain to algorithmically generate varied vertically challenging terrain for ACL.
    \item  We propose VertiSelector (VS) (Fig.~\ref{VS}), an ACL framework that samples training terrain based on estimates of future learning potential.
    \item Extensive simulation and hardware experiments demonstrate that our approach significantly enhances navigation performance compared against a manually designed curriculum, vanilla RL, a hybrid (classical and learning) method and two classical baseline approaches. 
\end{itemize}
\section{Related Work}
In this section, we review related work in off-road mobility using classical and data-driven methods, as well as curriculum learning techniques to improve learning methods. 

\subsection{Off-Road Mobility}
Off-road mobility is a challenging domain for autonomous robots, as they must navigate complex, unstructured environments with varying terrain conditions. Traditional approaches to off-road navigation often rely on hand-crafted perception~\cite{lu2014layered}, planning~\cite{rastgoftar2018data}, modeling~\cite{he2019review}, and control~\cite{williams2016aggressive} methods with human heuristics. Those classical methods require extensive engineering effort, are affected by cascading errors from upstream modules, and struggle at adapting to new environments~\cite{thrun2006stanley}. 

Considering the limitations of classical approaches, learning off-road mobility has emerged as a promising alternative avenue~\cite{xiao2022motion}, such as learning end-to-end policies~\cite{pan2020imitation}, semantic perception~\cite{ manduchi2005obstacle, maturana2018real, shaban2022semantic, meng2023terrainnet, viswanath2021offseg}, kinodynamic models~\cite{xiao2021learning, karnan2022vi, atreya2022high, datar2024terrain, pokhrel2024cahsor, maheshwari2023piaug}, parameter adaptation~\cite{xiao2020appld, wang2021appli, wang2021apple, xu2021applr, xiao2022appl}, and cost functions~\cite{sivaprakasam2021improving, dashora2022hybrid, cai2022risk, castro2023traversability, cai2024evora, seo2023learning, jung2024v, xiao2022learning}. 
Learning methods, such as RL, can alleviate engineering effort and allow emergent and adaptive behaviors. However, those methods are often data-hungry, either requiring extensive expert demonstration or labeled data for imitation learning or millions of trial-and-error exploration steps using RL. How to generalize learning results to unseen deployment enviornments is also difficult. 
To tackle those challenges of learning methods, curriculum learning has the potential to improve sample efficiency and generalization by presenting the robot with a sequence of tasks that gradually increase in difficulty. VS is based on RL guided by an efficient curriculum for wheeled mobility on vertically challenging terrain.  

\subsection{Curriculum Learning}
Curriculum learning is a concept inspired by the structured nature of human learning~\cite{elman1993learning}. This idea was expanded upon by Bengio et al.~\cite{bengio2009curriculum}, who proposed a learning paradigm where training examples are presented in a meaningful order, gradually increasing in complexity. Over the following years, curriculum learning found applications in various supervised learning settings, such as natural language processing~\cite{mikolov2013distributed} and computer vision~\cite{pentina2015curriculum}. Building on these foundational ideas, the community developed a set of mechanisms collectively known as Automatic Curriculum Learning (ACL)~\cite{portelas2020teacher}.

ACL techniques automatically adjust the distribution of training data by selecting learning situations that match the evolving capabilities of the learning agents. While ACL has been successfully applied to various domains, most applications have been limited to simple tasks or simulated environments. For instance, in supervised learning, ACL has been employed to improve performance on static benchmark datasets, such as image classification~\cite{bengio2009curriculum}. Similarly, in RL, ACL has been primarily studied in the context of simple gridworld environments~\cite{narvekar2020curriculum} and Atari games~\cite{justesen2018illuminating}.

Despite ACL's potential in improving sample efficiency and asymptotic performance in these simplified settings~\cite{andrychowicz2017hindsight}, few works have explored its application to real-world mobility tasks, where robots must learn to navigate complex, unstructured, off-road environments. The challenges posed by off-road terrain require sophisticated approaches to curriculum learning. In this work, we investigate how to automatically select an appropriate task sequence as a curriculum based on real-world data to enhance sample efficiency and policy generalization for wheeled robots navigating vertically challenging terrain, while considering their evolving capabilities. 

\section{Method}

In this section, we introduce the \texttt{VW-Chrono} simulator (Sec.~\ref{sec::vw_chrono}) and its corresponding PCG environments (Sec.~\ref{sec::pcg}). We also present our RL problem formulation for vertically challenging terrain in Sec.~\ref{sec::mdp} and sample efficient ACL framework, VertiSelector (VS), in Sec.~\ref{sec::vs}, which considers robot's future learning potential on different terrain.

\subsection{\texttt{VW-Chrono}}
\label{sec::vw_chrono}
To create a realistic simulation environment for vertically challenging terrain, we first collect elevation map data~\cite{miki2022elevation} using our physical Verti-4-Wheeler (V4W) on a custom-built indoor testbed. This testbed consists of hundreds of randomly distributed and stacked rocks and boulders, with an average size of 30cm, matching the scale of the V4W. The test course measures 3.1$\times$1.3m, with the highest elevation reaching up to 0.5m, more than twice the height of the vehicle (Fig.~\ref{VW-chrono} middle). We generate a $150 \times 150$ synthetic grayscale elevation map as part of state space to represent real terrain elevation distribution.

Within the Chrono multi-physics simulation engine, we generate a triangular mesh by assigning a vertex to each pixel of the elevation map. The mesh vertices are then vertically adjusted to align with the specified elevation values in the elevation map. Finally, the mesh is scaled to match the given spatial extents (Fig.~\ref{VW-chrono} around). To simulate the interaction between the terrain and the vehicle's wheels, we set the friction coefficient of the terrain material using Chrono's built-in material properties. The friction coefficient is set to 0.9, and the restitution coefficient is set to 0.01. These values are carefully chosen to represent a realistic off-road terrain surface, considering factors such as tire grip and surface deformation. By following this methodology, we create a highly accurate and realistic simulation environment that closely mimics the vertically challenging terrain encountered by the V4W in real-world scenarios.

\subsection{PCG Environments}
\label{sec::pcg}
The diversity of PCG environments makes them valuable testbeds for evaluating the robustness and generalization ability of RL agents. To maintain the PCG principle, we assign a fixed identifier (index) to each terrain. We generate a sequence of elevation maps by linearly interpolating between a starting map $I_0$ (flat terrain) and an ending map $I_N$ (real-world rugged terrain) using a weighted average. The intermediate map $I_i$ at index $i$ out of $N+1$ indices can be calculated using the following equation:
\begin{equation}
I_i = (1 - \frac{i}{N}) I_0 + \frac{i}{N} I_N,\quad\forall i\in\{0, 1, ..., N\}.
\label{eqn::PCG}
\end{equation}
The $N+1$ elevation maps generated by PCG serve as individual tasks that, when ordered appropriately, comprise our curriculum to learn wheeled mobility on vertically challenging terrain. 

\subsection{POMDP for Vertically Challenging Terrain}
\label{sec::mdp}

In this work, we formulate the off-road navigation task for wheeled robots as a Partially-Observable Markov Decision Process (POMDP) characterized by a tuple $(\mathcal{S}, \mathcal{A}, \mathcal{P}, \gamma, r, \mathcal{O}, \Omega)$, where $\mathcal{S}$ represents the complete state space, $\mathcal{A}$ denotes the action space, $\mathcal{P}: \mathcal{S} \times \mathcal{A} \to \mathbb{P}(\mathcal{S})$ signifies the transition probability function, $\gamma \in [0, 1)$ is the discount factor, $r: \mathcal{S} \times \mathcal{A} \to \mathbb{R}$ is the reward function, $\mathcal{O}$ is the observation space consisting of local terrain and current vehicle information, and $\Omega: \mathcal{S} \times \mathcal{A} \to \mathbb{P}(\mathcal{O})$ is the observation probability function that maps states and actions to observations. We employ RL to learn a policy $\pi: \mathcal{O} \to \mathcal{A}$  that maps observations $o \in \mathcal{O}$ to actions $a \in \mathcal{A}$, enabling the robot to navigate vertically challenging terrain while avoiding pitfalls such as getting stuck or rolling over, ultimately guiding it to reach the designated goal. The objective is to maximize the expected cumulative discounted reward:
\begin{equation}
J(\pi) = \mathbb{E}_{\tau \sim \pi}\left[\sum_{t=0}^\infty \gamma^t r(o_t, a_t)\right],
\label{eqn::objective}
\end{equation}
where $\tau = (o_0, a_0, o_1, a_1, \ldots)$ represents a trajectory sampled from the policy $\pi$.

To be specific, our observation space $\mathcal{O}$ includes the angular difference between the vehicle and goal heading, current vehicle velocity, and a low-dimensional representation of the elevation map patch underneath the robot obtained using a Sliced-Wasserstein Autoencoder (SWAE)~\cite{kolouri2018sliced}. The action space $\mathcal{A}$ consists of the desired linear speed and steering angle, which will be tracked by a low-level PID controller. We employ Proximal Policy Optimization (PPO)~\cite{schulman2017proximal} as the RL algorithm to learn the policy, considering the continuous action space.

The reward function $r$ is designed to incentivize the robot's progress toward the goal while penalizing immobilization due to excessive roll and pitch angles, as well as timeouts. It consists of three key components: (1) a progress reward that encourages the robot to move toward the goal, (2) an instability penalty that discourages excessive roll and pitch angles, and (3) a timeout penalty that penalizes the robot for not reaching the goal within a specified time limit.

By carefully designing the POMDP and reward function, we aim to learn a robust policy that enables a wheeled robot to navigate vertically challenging terrain efficiently and safely. The learned policy is then used as a foundation for our ACL approach, which further enhances RL sample efficiency and the robot's performance and generalization capabilities.

\subsection{VertiSelector (VS)}
\label{sec::vs}

VertiSelector (VS), depicted in Fig.~\ref{VS}, maintains a dynamic, non-parametric sampling distribution $q(\mathcal{T} | \mathcal{D}_\text{train})$ over the set of PCG-generated training terrain $\mathcal{D}_\text{train}$, favoring terrain with higher learning potential. Specifically, throughout the training process, VS updates $q(\mathcal{T} | \mathcal{D}_\text{train})$ according to a heuristic score that assigns greater weight to terrain with higher estimated learning potential based on the robot's past experiences. VS maintains two arrays, $\mathbf{u} \in \mathbb{R}^{|\mathcal{D}_\text{train}|}$ and $\mathbf{v} \in \mathbb{N}^{|\mathcal{D}_\text{train}|}$, where $u_i$ stores the score for terrain $\mathcal{T}_i$ and $v_i$ keeps track of the episode count at which $\mathcal{T}_i$ was last sampled. After each episode, VS updates $q(\mathcal{T} | \mathcal{D}_\text{train})$ by computing a mixture of two distributions: $q_u(\mathcal{T} | \mathcal{D}_\text{train})$, based on the terrain scores, and $q_v(\mathcal{T} | \mathcal{D}_\text{train})$, based on the elapsed time since each terrain was last sampled:
\begin{equation}
q(\mathcal{T} | \mathcal{D}_\text{train}) = (1 - \alpha) \cdot q_u(\mathcal{T} | \mathcal{D}_\text{train}) + \alpha \cdot q_v(\mathcal{T} | \mathcal{D}_\text{train}),
\label{eqn::mixture}
\end{equation}
where $\alpha \in [0, 1]$ is a hyperparameter regulating the equilibrium between the two distributions. The mixture distribution ensures that the sampling process considers both the estimated learning potential and the time elapsed since each terrain was last encountered, mitigating the risk of catastrophic forgetting in neural networks.

\subsubsection{Terrain Scoring Mechanism}

To gauge the learning potential of a terrain $\mathcal{T}_i$, VS assigns a score $u_i$ based on the robot's experience in the most recent episode on that terrain. The score is computed using the TD error $\delta_t = r(s_t, a_t) + \gamma V(s_{t+1}) - V(s_t)$, which quantifies the discrepancy between the expected and actual returns at each timestep. Higher-magnitude TD errors suggest a greater potential for learning from revisiting a particular state transition.
VS employs the Generalized Advantage Estimator (GAE) \cite{schulman2015high} to compute the terrain scores. The GAE at timestep $t$ is defined as:
\begin{equation}
\hat{A}_t = \sum_{k=t}^{T-1} (\gamma \lambda)^{k-t} \delta_k,
\label{eqn::gae}
\end{equation}
where $\lambda \in [0, 1]$ is a hyperparameter that governs the bias-variance trade-off of the advantage estimates and $T$ is the episode length. The terrain score $u_i$ is then computed as the average absolute value of the GAE over the episode:
\begin{equation}
u_i = \text{score}(\tau, \pi)= \frac{1}{T} \sum_{t=0}^{T-1} |\hat{A}_t|.
\label{eqn::score}
\end{equation}
The absolute value of the GAE is equal to the L1 loss between the estimated and true value functions, which is a suitable measure of the learning potential. Given the terrain scores, VS defines the score-based sampling distribution $q_u(\mathcal{T} | \mathcal{D}_\text{train})$ using a rank-based prioritization scheme:
\begin{equation}
q_u(\mathcal{T}_i | \mathcal{D}_\text{train}) = \frac{\text{rank}(u_i)^{-\beta}}{\sum_{j=1}^{|\mathcal{D}_\text{train}|} \text{rank}(u_j)^{-\beta}},
\label{eqn::score_dist}
\end{equation}
where $\text{rank}(u_i)$ is the rank of $u_i$ among all terrain scores in descending order, and $\beta > 0$ is a hyperparameter that allows us to tune how much $\text{rank}(u_i)$ ultimately determines the resulting distribution. This rank-based prioritization ensures that the sampling process focuses more on terrain with relatively higher learning potential while still maintaining some probability of selecting lower-scored terrain.

\subsubsection{Staleness-Aware Prioritization}

To prevent the terrain scores from becoming outdated and to encourage revisiting previously encountered terrain, VS incorporates a staleness-aware prioritization scheme. The staleness-based sampling distribution $q_v(\mathcal{T}_i | \mathcal{D}_\text{train})$ is defined as:
\begin{equation}
q_v(\mathcal{T}_i | \mathcal{D}_\text{train}) = \frac{n - v_i}{\sum_{j=1}^{|\mathcal{D}_\text{train}|} (n - v_j)},
\label{eqn::recency_dist}
\end{equation}
where $n$ is the total number of episodes sampled so far, and $v_i$ is the episode count at which terrain $\mathcal{T}_i$ was last sampled. This distribution assigns higher probability to terrain that have not been recently visited, encouraging the robot to update its knowledge of previously encountered terrain.

By combining the score-based and staleness-aware prioritization schemes, VS effectively balances the exploration of terrain with high learning potential and the exploitation of acquired knowledge, leading to more efficient and effective learning in vertically challenging environments.
\section{Implementation}

In this section, we present implementation details of the \texttt{VW-Chrono} simulator and the V4W physical robot.

\subsection{Simulation Setup}
An overview of the simulation environment is shown in
Fig.~\ref{TrainTest}. We generate 100 distinct synthetic terrain for training based on the real-world rock testbed~\cite{datar2023toward} and each terrain has a fixed identifier (index) by the PCG principle. To show the generalization of VS, the test terrain are more uneven than the training terrain, as visualized in Fig.~\ref{TrainTest}. We use a mobile robot with reduced double wishbone suspensions and rack-pinion steering in \texttt{VW-Chrono}. For RL training, an episode terminates if the robot reaches the designated goal or exceeds the maximum time (20s). To learn wheeled mobility on vertically challenging terrain, the following design choices are made:

\begin{figure}[h]
    \centering
    \includegraphics[width=\columnwidth]{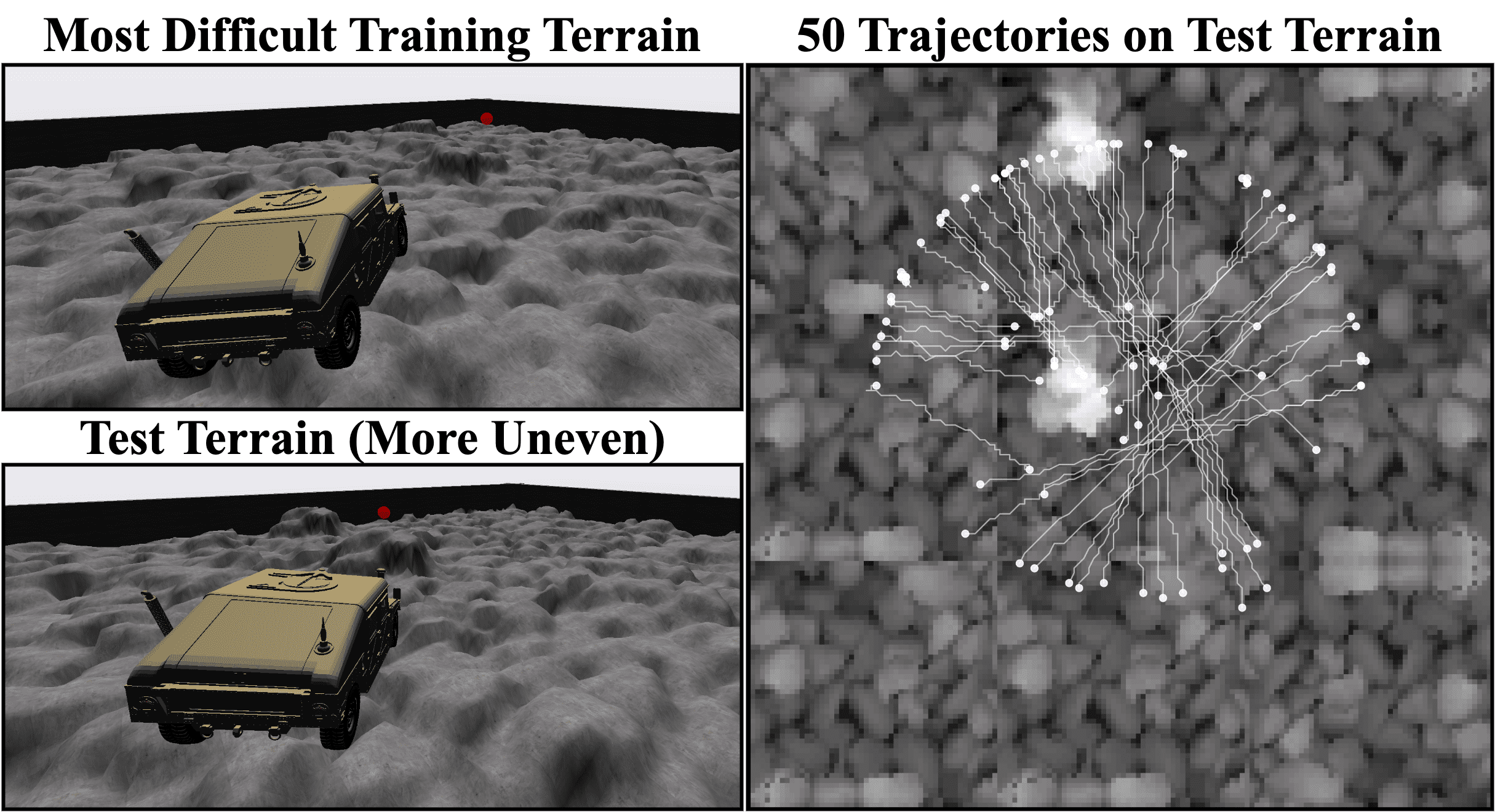}
    \caption{Overview of the Simulation Setup with the Most Difficult Training Terrain: The test terrain is more uneven than the training ones to evaluate the generalization of VS.} 
    \label{TrainTest}
\end{figure}

\subsubsection{Observation Space} 
The observation space consists of the components mentioned in Sec.~\ref{sec::mdp}. The angular difference between the vehicle and goal heading is normalized to $[-1, 1]$, and the vehicle velocity is normalized from $[-4, 4]$ m/s to $[-1, 1]$. The $64 \times 64$ elevation map is processed by the SWAE, which reduces it to a $64 \times 1$ latent vector. This vector is further compressed to a $16 \times 1$ vector using a fully connected neural network. Finally, the compressed elevation map is concatenated with two scalar values, resulting in an $18 \times 1$ state vector. 

\subsubsection{Action Space} 
The action space, as described in Sec.~\ref{sec::mdp}, includes the steering angle in the range of $[-1, 1]$ and the linear velocity normalized from $[-4, 4]$ m/s to $[-1, 1]$. A low-level PID controller generates appropriate throttle and steering commands based on these actions.

\subsubsection{Policy Architecture} 
We employ PPO~\cite{schulman2017proximal} with a policy network consisting of a shared feature extractor and separate fully-connected networks for the policy (actor) and value function (critic). The feature extractor takes the state space as input and passes it through a series of fully-connected layers with $\{64, 128, 64\}$ neurons and ReLU activations, outputting a 32-dimensional feature vector. Both the policy and value networks have two hidden layers with 64 neurons each and ReLU activations. 

\subsubsection{SWAE Architecture} 
The SWAE learns a compact representation of the elevation map. The encoder consists of convolutional layers with $\{32, 64, 128, 256, 512\}$ channels, kernel size 3, stride 2, padding 1, BatchNorm2d, and LeakyReLU activation, outputting a 64-dimensional latent vector mentioned above. The decoder mirrors the encoder's architecture in reversed order, with five convolutional transpose layers and channels decreasing from 512 to 32. The final output is a reconstructed $64 \times 64$ elevation map, passed through a Tanh activation. The SWAE is trained to minimize the Sliced Wasserstein Distance between the encoded latent space and a prior distribution.

\subsubsection{Reward Design}
The reward function for our RL agent consists of three main components mentioned in Sec.~\ref{sec::mdp}:
\begin{equation}
R_t = R_{\text{progress}} + R_{\text{rollover}} + R_{\text{timeout}}.
\end{equation}
The progress reward $R_\text{progress}$ incentivizes the robot to move toward the goal by providing positive rewards for the distance covered. A penalty is applied if the robot fails to move at least 1cm within 0.1 seconds:
\begin{equation}
R_\text{progress} = w_1 \cdot \Delta d - w_2 \cdot \mathbb{I}(\Delta d < 0.01),
\end{equation}
where $\Delta d$ is the distance moved toward the goal between the previous and current timestamps, $\mathbb{I}()$ is an indicator function, and $w_1$ and $w_2$ are weight terms.
To prevent rollovers, the rollover penalty $R_\text{rollover}$ discourages excessive roll and pitch angles:
\begin{equation}
R_{\text{rollover}} = -w_3 \cdot \sum_{i \in \{\text{roll}, \text{pitch}\}} \max(0, |\theta_i| - \alpha),
\end{equation}
where $\theta_{\text{roll}}$ and $\theta_{\text{pitch}}$ are the roll and pitch angles, respectively, $w_3$ is a weight term and $\alpha$ is a constant threshold angle. Finally, the timeout penalty $R_\text{timeout}$ is applied when the robot fails to reach the goal within a time limit $T$. It consists of a fixed penalty $c$ and an additional penalty based on the remaining distance to the goal:
\begin{equation}
R_\text{timeout} = -(w_4 \cdot d_\text{remain} + c) \cdot \mathbb{I}(t \geq T),
\end{equation}
where $d_{\text{remain}}$ is the remaining distance to the goal, $t$ is the current time, and $w_4$ is a weight term. Table \ref{tab::parameters} shows all hyper-parameters of our reward function.

\begin{table}[h]
\centering
\caption{Reward Weights}
\renewcommand{\arraystretch}{1.5}
\begin{tabular}{ccccccc}
\toprule[1pt]
$w_1$ & $w_2$ & $w_3$ & $w_4$ & $\alpha$ & $c$ & $T$ \\
\midrule
50 & 10 & 20 & 10 & 30 & 100 & 20 \\
\bottomrule
\label{tab::parameters}
\end{tabular}
\vspace{-20pt}
\end{table}

\subsection{V4W and Vertically Challenging Testbed}
To evaluate the performance of the learned policy, we deploy our model on the V4W platform ($0.863\text{m} \times 0.249\text{m} \times 0.2\text{m}$, Fig.~\ref{VW-chrono} middle), a four-wheeled vehicle based on an off-the-shelf, two-axle, four-wheel-drive, off-road platform from Traxxas. The onboard computation is handled by an NVIDIA Jetson Xavier NX module. A Microsoft Azure Kinect RGB-D camera produces depth images to construct real-time elevation maps~\cite{miki2022elevation}. We use low-gear and lock both front and rear differentials to improve mobility on vertically challenging terrain. For the controlled environment, we shuffle our indoor rock testbed to achieve varying levels of difficulty: easy, medium, and hard. The testbed is designed to mimic vertically challenging terrain encountered in outdoor off-road environments with controllable complexity.

\section{Experiments}

We present the simulation results in the \texttt{VW-Chrono} simulator and compare the performance of VS against other baselines designed for vertically challenging terrain.

\subsection{Baselines}
VS is compared against four baseline methods: Optimistic Planner (OP), Naive Planner (NP), Vanilla RL (VR), Manually-designed Curriculum (MC)~\cite{xu2024reinforcement} and WMVCT~\cite{datar2024learning}.

OP minimizes the angular difference between the vehicle's current and desired heading, optimistically assuming a flat terrain. However, this assumption often struggles with steep slopes and rugged boulders, leading to suboptimal performance on vertically challenging terrain. 

To address the limitations of OP, NP incorporates a heuristic based on the elevation map of the surrounding terrain. This planner divides the $64 \times 64$ surrounding elevation map into regions and selects the most traversable direction based on the mean and variance of the elevation values. Although more effective than OP, NP still relies on fixed rules and may not adapt well to diverse terrain conditions.

VR takes a more flexible approach by training the policy on randomly selected terrain from the PCG-produced training set, without any explicit curriculum. While this allows the robot to experience a wide range of terrain conditions, the lack of structure in the training process may lead to suboptimal sample efficiency, as the robot may spend too much time on terrain that is either too simple or too challenging for its current skill level. 

MC addresses this issue by following a curriculum that gradually increases the difficulty of the training terrain. The curriculum consists of five stages, each with a specific success rate threshold $\{1, 1, 0.8, 0.6\}$ that the robot must reach before progressing to the next one. However, MC may not always align with the robot's actual learning progress, potentially leading to inefficiencies in the training process.

WMVCT is a hybrid method based on a sampling-based motion planner and a decomposed 6-DoF vehicle-terrain dynamics model (bicycle model for x, y, and yaw, elevation map for z, and neural network prediction for roll and pitch). 
Despite its high efficiency, the decomposition also introduces inaccuracies compared against a full 6-DoF model.

\subsection{Simulation Results}
\label{sec::sim_res}

Our main findings are that (i) VS with rank prioritization ($\alpha=0.1, \beta=0.1$) significantly improves both sample efficiency during training and generalization on the test terrain, attaining the highest success rate in 50 trials out of all baselines evaluated; (ii) The relatively low average roll and pitch angles indicate that VS learns a stable policy that can effectively handle the uneven test terrain.

\subsubsection{Training Performance}
\begin{figure}[]
    \centering
    \includegraphics[width=\columnwidth]{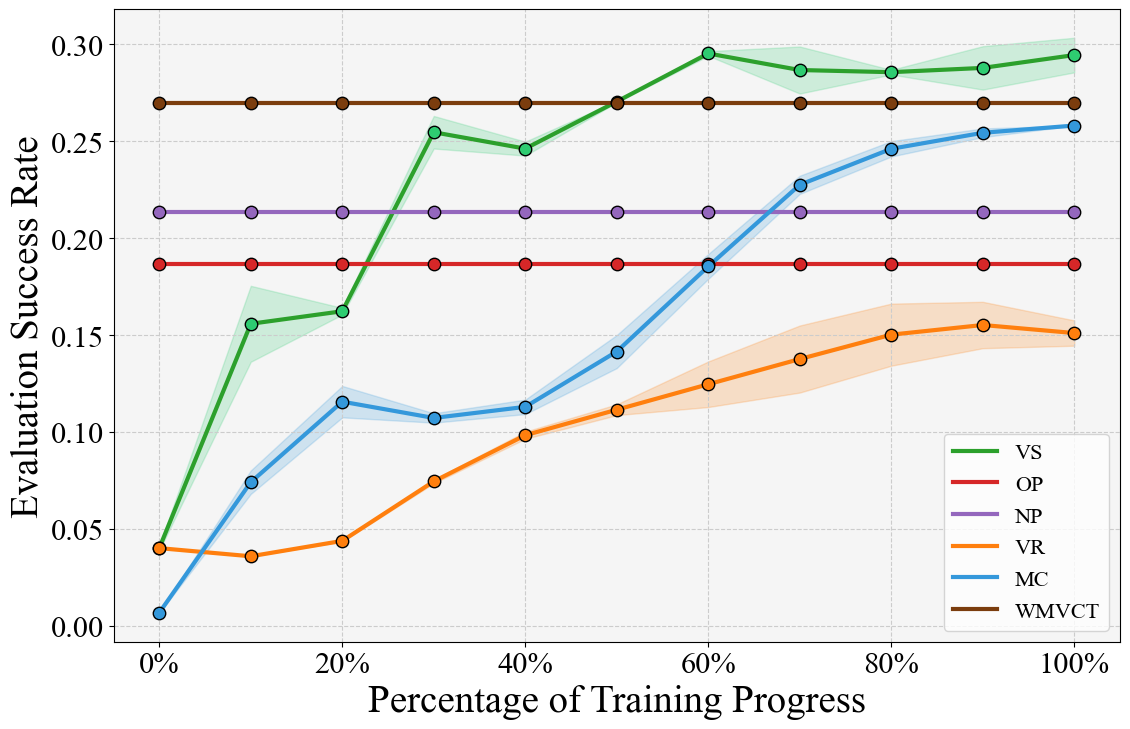}
    \caption{Smoothed Training Curves from Test Terrain Evaluation: VS is the most sample-efficient and generalizable. } 
    \label{RLCurve}
\end{figure}

As evident from the steeper slope of VS's training curve averaged over three runs in Fig.~\ref{RLCurve}, it consistently achieves higher evaluation success rates with fewer training samples compared to the baselines. This indicates that VS is more effective at extracting relevant information from the training terrain and updating its policy accordingly. The improved sample efficiency can be attributed to the automatic curriculum, which intelligently selects training terrain that is appropriately challenging for the robot's current skill level.

The training curve also demonstrates its superior generalization ability by converging to the highest success rate among all methods. By dynamically adjusting the difficulty, VS systematically exposes the robot to a diverse range of terrain features. This diversity helps the robot learn a more comprehensive and flexible policy that can better handle the variability and uncertainty of unseen environments. In contrast, the fixed heuristic of the planners (OP and NP), the decomposed 6-DoF vehicle-terrain dynamics model of WMVCT, the lack of a curriculum of VR, and the predefined curriculum of MC may overly specialize the robot to specific terrain types, limiting their generalization to novel test terrain.

\subsubsection{Evaluation Metrics}
The learned policies are evaluated on the test terrain using three metrics: 

\begin{enumerate}
    \item Number of successful trials (out of 50).
    \item Mean traversal time (of successful trials in seconds).
    \item Average roll/pitch angles with variance (in degrees).
\end{enumerate}

\begin{table}
\centering
\caption{Simulation Experiment Results}
\renewcommand{\arraystretch}{1.5}
\label{tab:eva_results}
\begin{tabular}{cccc}
\toprule[1pt]
Method & Success & Time & Angles (Roll/Pitch) \\
\midrule
\rowcolor[gray]{.9}
VS (Proposed) & \textbf{32/50} & 8.82$\pm$4.27 & \textbf{6.46$\pm$26.21} / 3.06$\pm$3.13 \\
MC & 26/50 & 8.46$\pm$5.19 & 6.55$\pm$17.56 / \textbf{2.8$\pm$1.08} \\
VR & 23/50 & 8.74$\pm$10.25 & 8.33$\pm$45.78 / 4.47$\pm$14.46 \\
WMVCT & 20/50 & 7.79$\pm$0.32 & 8.09$\pm$22.05 / 3.75$\pm$14.27 \\
OP & 14/50 & \textbf{7.56$\pm$0.37} & 10.09$\pm$50.1 / 4.59$\pm$13.13 \\
NP & 13/50 & 8.65$\pm$0.84 & 8.68$\pm$28.62 / 4.81$\pm$13.43 \\
\bottomrule
\end{tabular}
\end{table}

Table~\ref{tab:eva_results} summarizes the performance of the best model of each method on the test terrain and the best result for each metric is shown in bold. VS achieves the highest success rate, successfully navigating the test terrain in 32 out of 50 trials. This significantly outperforms all other baselines. In terms of traversal time, VS (8.82s) is comparable to OP (7.56s). However, OP's slightly faster time comes at the cost of a drastically lower success rate (14/50). VS strikes a balance between reliable navigation and reasonable speed. Moreover, VS maintains the lowest average roll (6.46$^{\circ}$) and second lowest pitch (3.06$^{\circ}$) angles within the 30$^{\circ}$ threshold, demonstrating its ability to keep the vehicle stable while traversing uneven terrain.

\subsection{Physical Results}
We conduct ten trials each on three configurations of our physical testbed (Fig.~\ref{outdoor_mobility} left) with the V4W, recording the same set of metrics. The results are presented in Table~\ref{tab:testbed_results}. The learned VS policy demonstrates a high success rate across all difficulty levels, with a slight decrease in performance as the terrain complexity increases. The average traversal time also shows a consistent trend, with longer times required for more challenging courses. MC fails more on the Medium course and fails all trials on the Hard one. It mostly suffers from longer traversal time and larger roll/pitch angles. These results validate VS's generalizability from simulation to a real-world vertically challenging testbed.

\begin{table}[h]
\centering
\caption{Physical Testbed Experiment Results}
\renewcommand{\arraystretch}{1.5}
\label{tab:testbed_results}
\begin{tabular}{ccccc}
\toprule[1pt]
Method & Difficulty & Success & Time & Angles (Roll/Pitch) \\
\midrule
\multirow{3}{*}{VS (Proposed)} & Easy & 8/10 & \cellcolor[gray]{.9}\textbf{13.99} & \cellcolor[gray]{.9}\textbf{0.15/0.53} \\
& Medium & \cellcolor[gray]{.9}\textbf{7/10} & \cellcolor[gray]{.9}\textbf{15.85} & \textbf{2.05}/1.85 \\
& Hard  & \cellcolor[gray]{.9}\textbf{5/10} & \cellcolor[gray]{.9}\textbf{20.86} & \textbf{0.25}/8.03 \\
\midrule
\multirow{3}{*}{MC} & Easy & \cellcolor[gray]{.9}\textbf{9/10} & 18.07 & 3.63/0.92 \\
& Medium & 6/10 & 17.22 & 3.97/\textbf{1.49} \\
& Hard  & 0/10 & N/A & 6.32/\textbf{0.9} \\
\bottomrule
\end{tabular}
\end{table}

\subsection{Outdoor Demonstration}
\begin{figure}[h]
\centering
\includegraphics[width=\columnwidth]{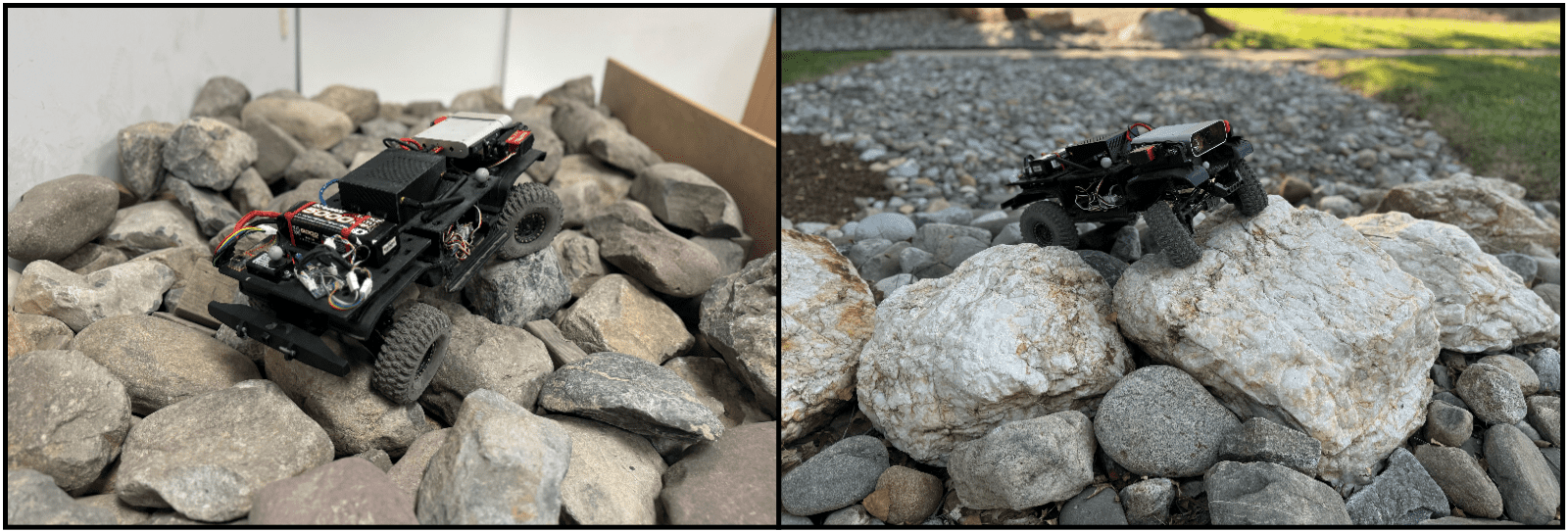}
\caption{Indoor Experiments and Outdoor Demonstration.}
\label{outdoor_mobility}
\end{figure}

To further demonstrate the generalizability and applicability of the learned VS policy, we deploy it on the V4W in a real-world outdoor environment. We select a challenging off-road location with diverse terrain features, including steep slopes, various rocks, and uneven surfaces (Fig.~\ref{outdoor_mobility} right). 
% The V4W is tasked with navigating from a starting point to a designated goal location using the learned policy.
% Fig.~\ref{outdoor_mobility} presents a snapshot of the V4W successfully traversing the outdoor terrain. 
The platform exhibits stable and efficient navigation by effectively making appropriate steering and throttle decisions based on the perceived outdoor terrain features. 
% The learned policy, guided by VS, enables the V4W to make appropriate steering and throttle decisions based on the perceived outdoor terrain features. 
% The outdoor demonstration shows VS's real-world applicability and generalization capability to enable wheeled robots to navigate vertically challenging terrain.

\section{Conclusions and Limitations}

This work introduces VS, an automatic curriculum learning framework that enhances sample efficiency and generalization of reinforcement learning for wheeled robot navigation on vertically challenging terrain. VS selectively samples training terrain based on the robot's evolving capabilities and learning potential to accelerate learning and facilitate robust navigation. The \texttt{VW-Chrono} simulator enables the generation of diverse and challenging terrain for training and testing. Simulation experiments demonstrate VS's superior performance compared to baseline methods, achieving a 23.08\% improvement in success rate. The real-world applicability of VS is validated through successful deployment on the physical V4W platform in both an indoor testbed and outdoor environment.

One of the limitations of VS is its focus on a specific set of terrain geometry and vehicle configurations, which may limit its generalizability to more diverse off-road scenarios. Despite the effectiveness of \texttt{VW-Chrono}, sim2real gap still remains: the framework trains and evaluates using a simulated full-scale vehicle, while real-world validation employs a scaled RC car. This discrepancy in vehicle scale and dynamics may undermine the transferability of results to practical scenarios. And it is also important to note that real rocky terrain exhibits spatially varying friction and restitution properties. Future work can focus on extending VS to incorporate various terrain semantics and vehicle types, as well as integrating with other learning paradigms such as imitation learning.

\section*{Acknowledgments}
This work has taken place in the RobotiXX Laboratory at George Mason University. RobotiXX research is supported by National Science Foundation (NSF, 2350352), Army Research Office (ARO, W911NF2320004, W911NF2420027, W911NF2520011), Air Force Research Laboratory (AFRL), US Air Forces Central (AFCENT), Google DeepMind (GDM), Clearpath Robotics, Raytheon Technologies (RTX), Tangenta, Mason Innovation Exchange (MIX), and Walmart.


%%%%%%%%%%%%%%%%%%%%%%%%%%%%%%%%%%%%%%%%%%%%%%%%%%%%%%%%%%%%%%%%%%%%%%%%%%%%%%%%

%%%%%%%%%%%%%%%%%%%%%%%%%%%%%%%%%%%%%%%%%%%%%%%%%%%%%%%%%%%%%%%%%%%%%%%%%%%%%%%%
\bibliographystyle{IEEEtran}
\bibliography{IEEEabrv,references}

% \begin{thebibliography}{99}

% \bibitem{c1} M. D. Teji, T. Zou, and D. S. Zeleke, “A survey of off-road mobile robots: Slippage estimation, robot control, and sensing technology,” \textit{Journal of Intelligent \& Robotic Systems}, vol. 109, p. 38, Oct 2023.
% \bibitem{c2} J. Schulman, F. Wolski, P. Dhariwal, A. Radford, and O. Klimov, “Proximal policy optimization algorithms,” \textit{arXiv preprint arXiv:1707.06347}, 2017.
% \bibitem{c3} A. Tasora, R. Serban, H. Mazhar \textit{et al}, “Chrono: An open source multi-physics dynamics engine,” in \textit{High Performance Computing in Science and Engineering: Second International Conference}, pp. 19-49, 2015.
% \bibitem{c4} A. Datar, C. Pan, M. Nazeri and X. Xiao, “Toward wheeled mobility on vertically challenging terrain: Platforms, datasets, and algorithms,” \textit{arXiv preprint arXiv:2303.00998}, 2023.
% \bibitem{c5} S. Narvekar, B. Peng, M. Leonetti, J. Sinapov, M. E. Taylor and P. Stone, “Curriculum learning for reinforcement learning domains: A framework and survey,”. \textit{Journal of Machine Learning Research}, pp. 1-50, 2020.
% \bibitem{c6} S. Kolouri, C. E. Martin, and G. K. Rohde, “Sliced-wasserstein autoencoder: an embarrassingly simple generative model,” \textit{arXiv.1804.01947}, 2018.

% \end{thebibliography}

\end{document}